\documentclass[letterpaper, 10 pt, conference]{ieeeconf}  

\IEEEoverridecommandlockouts                              

\usepackage[utf8]{inputenc} 
\usepackage[T1]{fontenc}    
\usepackage{hyperref}       
\usepackage{url}            
\usepackage{booktabs}       
\usepackage{amsfonts}       
\usepackage{nicefrac}       
\usepackage{microtype}      
\usepackage{xcolor}         
\usepackage{graphicx}
\usepackage{hyperref}
\usepackage{multirow}
\usepackage{dsfont}
\usepackage{amsmath}
\usepackage{amssymb}
\usepackage{algorithm}
\usepackage{algorithmic}
\usepackage[noadjust]{cite}
\usepackage{caption}

\DeclareMathOperator*{\argmax}{arg\,max}

\newcommand{\rev}[1]{\textcolor{black}{#1}}

\title{\LARGE \bf
NOPA: Neurally-guided Online Probabilistic Assistance\\for Building Socially Intelligent Home Assistants
}

\author{Xavier Puig$^{*}$, Tianmin Shu$^{*}$, Joshua B. Tenenbaum, Antonio Torralba
  \thanks{$^{*}$ Equal contribution. All authors are with MIT. \texttt{\{xpuig, tshu, jbt, torralba\}@mit.edu}}
}

\begin{document}


\maketitle
\thispagestyle{empty}
\pagestyle{empty}

\begin{abstract}
In this work, we study how to build socially intelligent robots to assist people in their homes. In particular, we focus on assistance with online goal inference, where robots must simultaneously infer humans' goals and how to help them achieve those goals. Prior assistance methods either lack the adaptivity to adjust helping strategies (i.e., when and how to help) in response to uncertainty about goals or the scalability to conduct fast inference in a large goal space. Our NOPA (Neurally-guided Online Probabilistic Assistance) method addresses both of these challenges. NOPA consists of (1) an online goal inference module combining neural goal proposals with inverse planning and particle filtering for robust inference under uncertainty, and (2) a helping planner that discovers valuable subgoals to help with and is aware of the uncertainty in goal inference. We compare NOPA against multiple baselines in a new embodied AI assistance challenge: Online Watch-And-Help, in which a helper agent needs to simultaneously watch a main agent's action, infer its goal, and help perform a common household task faster in realistic virtual home environments. Experiments show that our helper agent robustly updates its goal inference and adapts its helping plans to the changing level of uncertainty.\footnote{Code is available at \url{https://github.com/
xavierpuigf/online_watch_and_help}. Project website: \url{https://www.tshu.io/online_watch_and_help}.}
\end{abstract}

\section{Introduction}

There has been growing interest in engineering socially intelligent robots that can safely and productively work with humans in the real world. Prior work on robot assistance has achieved some success in scenarios where robots are given the true human goals a priori or only need to help humans in simple environments with a small state space. However, it remains very challenging to build robot assistants that can help humans perform all the activities of daily life in more natural settings, such as in our homes, where the space of human goals is vast and a person's goal at any point in time will not generally be known with certainty. 

\begin{figure}
    \centering
    \includegraphics[width=1.0\linewidth]{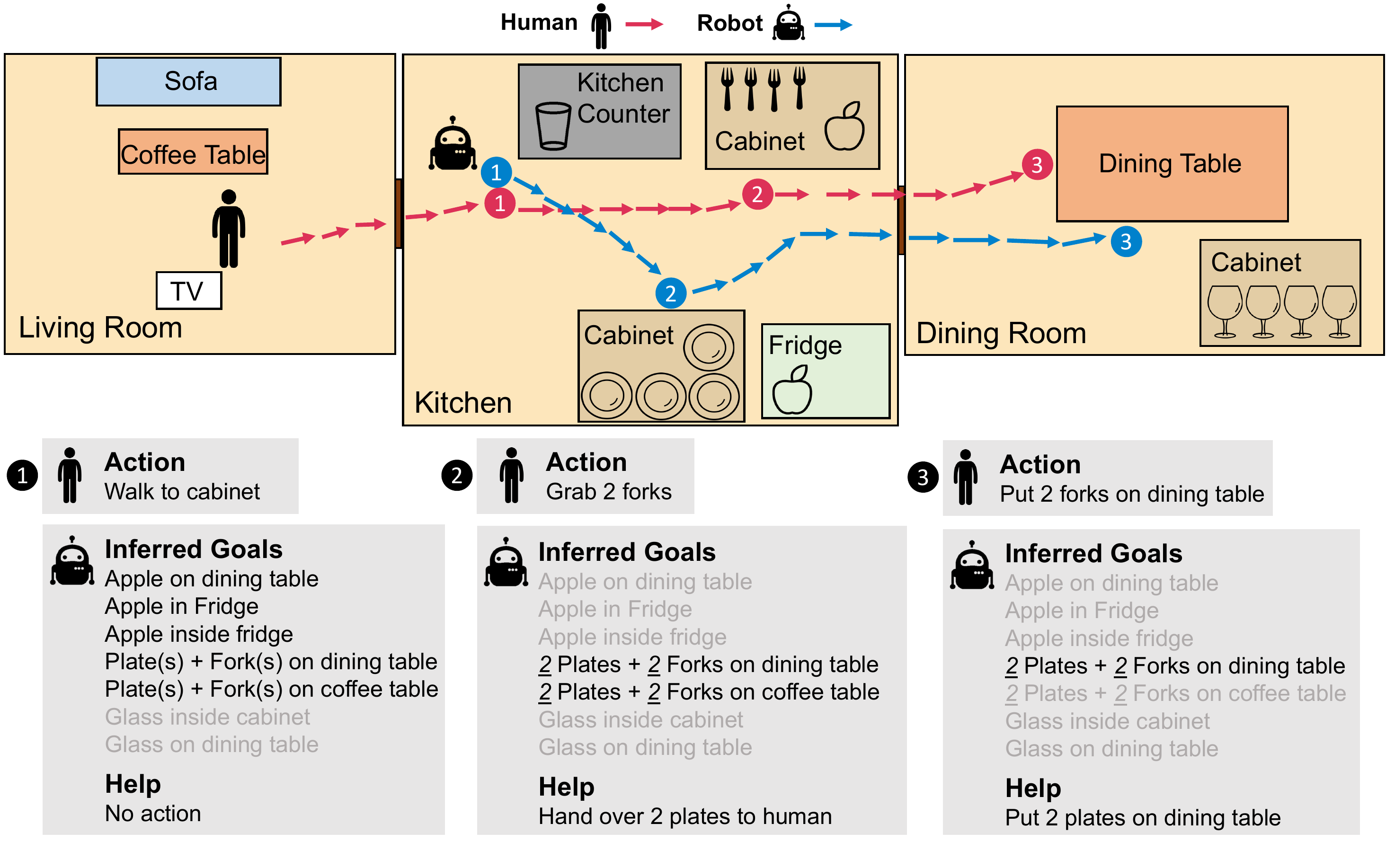}
    \caption{Illustration of successful online assistance. The robot initially has no knowledge about the human's goal and thus would opt to observe. As it observes more human actions, it becomes more and more confident in its goal inference, so it would dynamically adjust its helping subgoal. For instance, in this figure, the robot first sees the human walking towards a cabinet and consequently infers that the goal involves objects inside the cabinet. After the human grabs 2 forks, the robot infers that the goal is to put 2 sets of dining pieces (plates and forks) on the dining table or the coffee table but is uncertain about the goal location. Thus, it hands over 2 plates to the human instead of randomly guessing a location.}
    \label{fig:cover_fig}
\end{figure}

Our goal here is to build robot assistants that are able to help people perform a wide range of tasks in complex home environments. Our robot assistants must have the ability to infer the true goals of humans based on past observations in an online fashion, plan how to help humans without disrupting them, and adapt to their behaviors by simultaneously updating goal inference and helping strategies as the task progresses (as illustrated in Fig.~\ref{fig:cover_fig}). Such ability has proven difficult for robots to date due to two main technical obstacles. On the one hand, online goal inference in realistic environments is extremely difficult due to large state, action, and goal spaces; on the other hand, inaccurate or ambiguous goal inferences often lead to ineffective or even counterproductive attempts to help in systems that are not aware of their own uncertainty. 

\begin{figure*}[t!]
\centering
\includegraphics[width=1.0\textwidth]{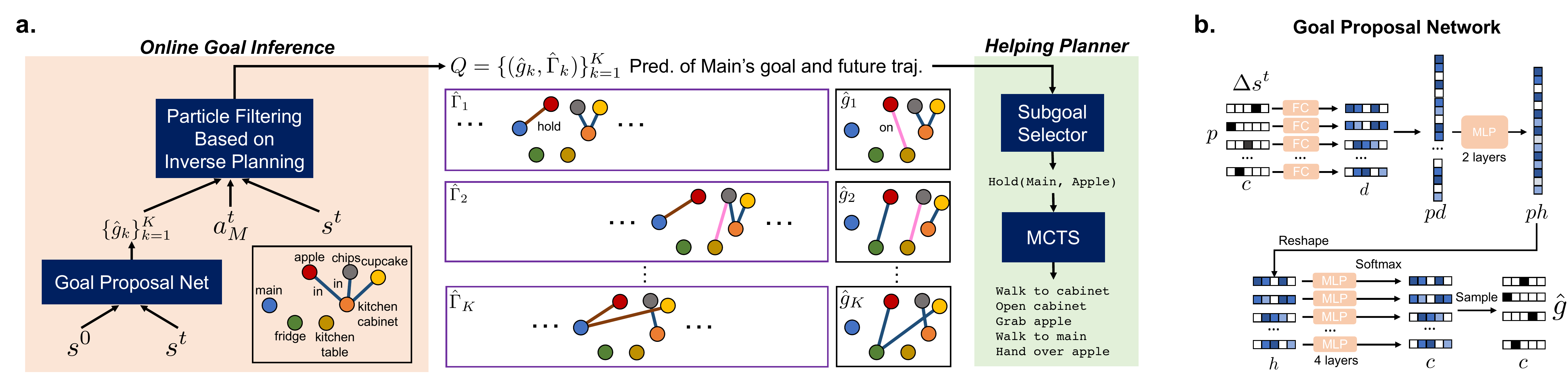}
\caption{(\textbf{a}) Overview of our approach, which consists of an online goal inference module and a helping planner. We represent states and goals using scene graphs. $s^t$ is the state at time $t$;  $a^t_\text{M}$ is the main agent's action at time $t$; $\hat{g}_k$ is the $k$-th goal proposal; and $\hat{\Gamma}_k$ is the prediction of the main agent's future trajectory corresponding to $\hat{g}_k$. (\textbf{b}) Goal proposal network. $\Delta s^t$ is a matrix encoding the difference between the predicate counts in the states $s^t$ and $s^0$, where each row is a one-hot vector, indicating the change in the count of a specific predicate. $p=136$ is the number of all predicate types, $c=9$ is the maximum number of counts, and $d=100$ and $h=128$ are the dimensions of the intermediate layers.}
\label{fig:overview}
\end{figure*}

To address these challenges, we propose a novel online assistance method, NOPA (Neurally-guided Online Probabilistic Assistance). As illustrated in Fig.~\ref{fig:overview}\textbf{a}, NOPA consists of two main components: (1) a \rev{neurally-guided} online goal inference module and (2) an uncertainty-aware helping planner. The \rev{neurally-guided} online goal inference module first produces bottom-up goal proposals from a neural \rev{network and} then maintains a set of predictions of goals and future trajectories consistent with the observed actions via particle filtering and inverse planning. This ensures that inferences are both fast and robust. Given the latest predictions and their certainty, the helping planner first identifies a subgoal that is most valuable to help with and then plans the corresponding helping actions using a symbolic planner. The resulting helping plan can adapt to all levels of uncertainty in the predictions. For instance, when there are multiple possible target locations for a goal object, the robot assistant will deliver the object to the human agent instead of risking misplacing the object.

For evaluation, we present a new embodied AI assistance challenge, Online Watch-And-Help (O-WAH), which builds off a recent multi-agent virtual home environment, VirtualHome-Social. Unlike most existing challenges (e.g., \cite{carroll2019utility,puig2021watchandhelp}), in O-WAH, a helper agent needs to infer the goal of a main agent in an online fashion and simultaneously help achieve the inferred goal as efficiently as possible. We evaluate agents built with NOPA and several baselines in a range of household tasks, helping the main agent controlled by either a human player or a planner-based agent. The experimental results show that NOPA significantly outperforms all baselines. We are also able to observe intelligent helping strategies emerging from NOPA adapting to new observations.

In sum, our main contribution includes (1) a new online assistance method for building socially intelligent home assistants in complex settings and (2) an embodied AI assistance challenge, Online Watch-And-Help, as a testbed for training and testing embodied agents to perform online goal inference and helping in realistic virtual home environments.

\section{Related Work}
\textbf{Online goal inference.} Online goal inference approaches generally fall into two categories -- (1) feedforward prediction that directly maps observed past trajectories to possible goals, typically enabled by goal prediction networks \cite{rabinowitz2018machine,cao2020long,liu2020spatiotemporal,nan2020learning,dendorfer2020goal,zhao2020tnt,yao2021bitrap, tran2021goal,mangalam2021goals}, or (2) generative approaches such as inverse planning \cite{carberry2001techniques,ramirez2009plan,sohrabi2016plan,baker2017rational,ullman2009help,zhi2020online,netanyahu2021phase,tejwani2022social} which conduct inference by comparing generated plans of given goal hypotheses with observed actions. Feedforward methods are fast and can perform well in simple tasks (such as destination prediction for pedestrians \cite{liu2020spatiotemporal}) when trained with a large amount of data. However, in unfamiliar scenarios, inverse planning methods often outperform feedfoward prediction due to their ability to imagine rational behaviors under various conditions. One of the limitations of inverse planning methods is that they can be slow if the goal space is large and rely on manually designed heuristics to speed up the inference \cite{zhi2020online,netanyahu2021phase}. Our work integrates these two types of approaches to achieve both speed and robustness.

\textbf{Embodied AI assistance with unknown goals.} There has been a rich history of research on embodied AI assistance. Many of the existing works assume a known common goal shared among human and AI/robot partners \cite{goodrich2007human,lasota2017survey,nikolaidis2015efficient,rozo2016learning,jaderberg2019human,carroll2019utility,bard2020hanabi,hu2020other,wu2021too}. However, in the real world, robots often need to infer humans' goals on the fly. There has been work on helping with inferred goals \rev{\cite{fern2014decision,liu2016goal,albrecht2018autonomous,hadfield2016cooperative,javdani2018shared,puig2021watchandhelp}} which shows that an accurate goal inference can improve the objective and perceived performance of embodied AI assistants. However, when the goal inference is uncertain, helping with inferred goals often leads to counterproductive behaviors such as undoing finished goals \cite{puig2021watchandhelp}. For this, some prior work devised planners under uncertain goal inference in simple environments \cite{shah2020planning}. A recent study proposed a goal-agnostic assistance framework via empowerment \cite{du2020ave}, which aims at changing the states to maximize an agent's ability to reach as many goals as possible regardless of its true intent. Despite its success in certain domains, assisting humans in real-world settings without the knowledge of their goals would often result in counterproductive behaviors. \rev{Our work investigates how to design an uncertainty-aware planner that intelligently adjusts the helping behavior ranging from goal-agnostic strategies to goal-specific plans in a complex environment.}

\textbf{Embodied AI assistance benchmarks.} There have been benchmarks designed to evaluate AI agents' ability to collaborate with human teammates \cite{carroll2019utility,bard2020hanabi}. However, most of them focus on simple game environments and assume a common goal given to both the AI and human agents a priori. Based on a realistic virtual home environment, \cite{puig2021watchandhelp} proposed a challenge, Watch-And-Help, in which a helper agent must infer a main agent's goal from a pre-recorded demonstration, and then help the main agent to achieve the same goal. We extend this challenge to an online assistance setting, where the demonstration is no longer available. As a result, helper agents have to pay attention to the main agent's actions and constantly update the inference and its uncertainty while working towards inferred goals, which is closer to what robots are expected to do in real homes. Such benchmark complements conventional robot assistance studies conducted in lab environments \cite{goodrich2007human,lasota2017survey}, providing a reproducible and scalable way to compare different methods.

\section{Method}

\subsection{Problem Setup}\label{sec:problem_setup}

We define the online assistance problem as a mixed-observability Markov decision process (MOMDP) \cite{ong2010planning}, where a \textit{\textbf{helper}} agent needs to infer a \textit{\textbf{main}} agent' goal and help the main agent achieve its goal faster. This can be formalized by $\langle \mathcal{S}, \mathcal{G}, \mathcal{A}_\text{H}, \mathcal{O}, \mathcal{T}_\mathcal{S}, \mathcal{T}_\mathcal{G}, Z, R_\text{H}, \gamma \rangle$. The overall state has two components: the world state, $s \in \mathcal{S}$, which is fully observable to both agents, and the main agent's goal, $g \in \mathcal{G}$, which is partially observable to the helper agent. $\mathcal{A}_\text{H}$ is the action space of the helper agent. The helper agent's observation consists of the world state and the main agent's action, i.e., $\mathcal{O} = \mathcal{S} \times \mathcal{A}_\text{M}$. $\mathcal{T}_\mathcal{S}(s, g, a_\text{H}, s^\prime)=p(s^\prime | s, g,  a_\text{H})$ is the transition function for the world state, and $\mathcal{T}_\mathcal{G}(s, g, a_\text{H}, s^\prime, g^\prime) = p(g^\prime | s, g, a_\text{H}, s^\prime)$ is the transition function for the goal. $Z(s^\prime, g^\prime, a_\text{H}, o) = p(o=(s, a_\text{M}) | s^\prime, g^\prime, a_\text{H})$ is the conditional probability function for the observation result. At step $t+1$, the helper infers the main's goal given main's past trajectory upon $t$, i.e., $\Gamma^t = \{(s^\tau, a_\text{M}^\tau)\}_{\tau=1}^t$. The expected reward function for the helper agent is defined as $R_\text{H}(s,a|\Gamma^t)=E_{p(g|\Gamma^t)}[\mathds{1}(s=g)] - c_\text{H}(a)$, where $c_\text{H}(a)$ is the cost for action $a$, and $\mathds{1}(\cdot)$ checks if the goal is satisfied in the current world state $s$. $\gamma$ is the discount factor. Note that the assumption that the world states are fully observable for both agents is common in prior work on assistance with unknown goals \cite{liu2016goal,hadfield2016cooperative,javdani2018shared}.

\subsection{Method Overview}
\label{sec:method_overview}

\begin{algorithm}[t!]
\footnotesize
   \caption{NOPA}
   \label{alg:NOPA}
\begin{algorithmic}[1]
   \STATE {\bfseries Input:} $\Gamma_\text{M}^0 = \{(s^0, a^0_\text{M})\}$, $s^t$, $K$, $T_\text{max}$, $T_\text{prop}$, $q$, $w_r$, $w_c$, $w_m$, $L_\text{max}$
   \STATE $t \leftarrow 1$, $l \leftarrow 0$
   \STATE $Q \leftarrow \emptyset$
   \REPEAT 
   \STATE $Q, l \leftarrow \textbf{GoalInf}(t, Q, \Gamma_\text{M}^{t-1}, s^t, q, K, l, T_\text{prop})$
   \STATE $\Gamma_\text{H}^t \leftarrow \textbf{HelpPlanner}(Q, s^0, s^t, w_r, w_c, w_m, L_\text{max})$
   \STATE Execute the first action from the helping plan $a_\text{H}^t$
   \STATE Observe $a^t_\text{M}, s^{t+1}$ from the environment
    \STATE $t \leftarrow t + 1$
   \UNTIL $t = T_\text{max}$ or the true goal has not been reached
\end{algorithmic}
\end{algorithm}


To solve the online assistance problem formalized above, we propose NOPA (Neurally-guided Online Probabilistic assistance). Fig.~\ref{fig:overview}\textbf{a} provides an overview of NOPA, showing the two main components: i) Neurally-guided online goal inference, and ii) an uncertainty-aware helping planner. As sketched in Algorithm~\ref{alg:NOPA}, NOPA updates a set of particles conditioned on observed states and the main agent's actions. Each particle corresponds to a possible final goal. Common assistance frameworks \cite{liu2016goal,hadfield2016cooperative,javdani2018shared,puig2021watchandhelp} typically only consider the final goal for helping. However, when there is uncertainty in the goal inference, an intelligent assistant should seek intermediate subgoals that can be helpful with high certainty. For that, we also predict the main agent's future trajectory for each particle. \rev{We represent both intermediate states and final goals as a set of edges in a scene graph \cite{johnson2015image,liao2019synthesizing}, $\langle O, E \rangle$, as shown in Fig.~\ref{fig:overview}\textbf{a}. Each node, $o \in O$, represents an entity (agent/object); each edge, $e \in E$, corresponds to a predicate (e.g., \texttt{IN(apple, kitchencabinet)}), indicating the spatial relationship between two entities. Such representations have been widely adopted in robotics and embodied AI \cite{yan2020robotic, nguyen2020self, zhu2021hierarchical, srivastava2022behavior}.} Given the particles, the helping planner assesses the value of the edges in the intermediate states and the final goals and selects the most valuable edge as the helping subgoal. We introduce the details below.

\begin{algorithm}[t!]
\footnotesize
   \caption{GoalInf}
   \label{alg:online_goal_inference}
\begin{algorithmic}[1]
   \STATE {\bfseries Input:} $t$, $Q$, $\Gamma_\text{M}^{t-1} = \{(s^\tau, a^\tau_\text{M})\}_{\tau=0}^{t-1}$, $s^t$, $q$, $K$, $l$, $T_\text{prop}$
   \STATE {\bfseries Output:} Updated proposals $Q^\prime$ and steps since last proposal $l^\prime$
   \STATE $Q^\prime \leftarrow \emptyset$
   \IF {$Q \neq \emptyset$ and $l < T_\text{prop}$} 
        \FOR {$k=1,\cdots,|Q|$}
        \IF {$a^{t-1}_\text{M}$ is part of the plan $\hat{\Gamma}_k$}
            \STATE $Q^\prime \leftarrow Q^\prime \cup \{(\hat{g}_k, \hat{\Gamma}_k)\}$
        \ENDIF
        \ENDFOR
   \ENDIF
   \IF {$Q^\prime = \emptyset$}
    \FOR {$k=1,\cdots,K$}
        \STATE $\hat{g}_k \sim q(g | s^0, s^t)$ // Sample a goal proposal
        \STATE $ \hat{\Gamma}_k \leftarrow \textbf{MCTS}(s^t, \hat{g}_k, T_\text{prop})$ // Sample a plan
        \STATE $Q^\prime \leftarrow Q^\prime \cup \{(\hat{g}_k, \hat{\Gamma}_k)\}$
    \ENDFOR
    \rev{\STATE $l^\prime \leftarrow 0$ }
   \ELSE
    \STATE $l^\prime \leftarrow l + 1$    
   \ENDIF
   
   \RETURN {$Q^\prime$, $l^\prime$}
\end{algorithmic}
\end{algorithm}

\subsection{\rev{Neurally-guided} Online Goal Inference}

Unlike prior work on online goal inference, the objective of the online goal inference in this work is to help the downstream task, i.e., assistance. This poses additional challenges: (1) the helper agent has to estimate the uncertainty in the inference instead of only predicting the most probable goal; (2) it has to ensure that the inference is resilient in a dynamic environment; and (3) the inference has to be efficient so that the helper can have a prompt reaction to offer assistance.  For this, we propose a \rev{neurally-guided} online goal inference algorithm as summarized in Algorithm~\ref{alg:online_goal_inference}, which combines inverse planning and a neural network.

We use a goal proposal network (GPN), as depicted in Fig.~\ref{fig:overview}\textbf{b}, to learn a proposal distribution $q(g | s^0, s^t)$, from which we can sample $K$ goal proposals $\{\hat{g}_k\}_{k=1}^K$ given the initial state $s^0$ and the current state $s^t$. Each goal proposal is a set of goal predicates. Here, we only consider the first state and the current state instead of a sequence of past states for the input to GPN so that the GPN trained on episodes with only the main agent performing the tasks can be robustly applied to the helping condition where the sequence of state changes could become very different from the training sequences.

We use inverse planning to evaluate the goals proposed by the GPN and reject the ones that are inconsistent with the observed main agent's actions. To model \rev{an agent's behavior given each goal $\hat{g}_k$ with bounded rationality}, we use \rev{the built-in planner} to predict the future trajectory of the main agent in the next $T_\text{prop}$ steps $\hat{\Gamma}_k=\{(\hat{s}^{t+\tau_k},\hat{a}^{t+\tau_k})\}_{\tau=1}^{T_\text{prop}}$. \rev{Specifically, the level of rationality can be adjusted by the number of simulations and the length of rollouts}. We then create $K$ particles $Q = \{\hat{g}_k, \hat{\Gamma}_k\}$. Whenever the main agent takes a new action, $a_\text{M}^t$, we check if it is part of the predicted plan for each particle. If for a particle $k$, the action is not included in the predicted plan, then it suggests that the rational behavior under the corresponding goal is not consistent with the observed action. Thus the goal is likely to be wrong and the particle needs to be rejected. When there is no particle left or we have reached the prediction horizon $T_\text{prop}$, we resample another $K$ goals from the GPN based on the latest state and create new particles. Since some particles may share the same goal but have different predicted plans, our approach can consider different ways to reach a goal.

\begin{algorithm}[t!]
\footnotesize
   \caption{HelpPlanner}
   \label{alg:helping_planner}
\begin{algorithmic}[1]
   \STATE {\bfseries Input:} $Q$, $s^0$, $s^t$, $w_r$, $w_c$, $w_m$, $L_\text{max}$
   \STATE {\bfseries Output:} helping plan $\Gamma_\text{H}^t$
    \FOR {$e \in \mathcal{E}$}
        \STATE $L_\text{M}(e) \leftarrow \infty$
        \STATE $p(e) \leftarrow 0$
        \STATE $V(e) \leftarrow -\infty$
        \STATE $\Gamma_\text{H}(e) \leftarrow \emptyset$
        \IF {$e$ appears in the initial state}
            \STATE $L_\text{M}(e) \leftarrow 0$
            \STATE $p(e) \leftarrow 1$
        \ELSE
            \FOR {$k=1,\cdots,|Q|$}
                \IF {$e$ appears in the future traj. $\hat{\Gamma}_k$}
                    \STATE Let $\tau_k(e)$ be the first step when $e$ appears in $\hat{\Gamma}_k$
                    \STATE $L_\text{M}(e) \leftarrow \min(L_\text{M}(e), \tau_k(e))$
                    \STATE $p(e) \leftarrow p(e) + 1/|Q|$
                \ELSIF {$e$ appears in the predicted goal $\hat{g}_k$}
                    \STATE $L_\text{M}(e) \leftarrow \min(L_\text{M}(e), L_\text{max})$
                    \STATE $p(e) \leftarrow p(e) + 1/|Q|$
                \ENDIF
            \ENDFOR
        \ENDIF
        \IF {$p(e) > 0$}
            \STATE $\Gamma_\text{H}(e) \leftarrow \textbf{MCTS}(s^t, \{e\}, \infty)$ // Helper's plan for subgoal $e$
            \STATE $L_\text{H}(e) \leftarrow |\Gamma_\text{H}(e)|$
            \STATE Compute $V(s^t, e)$ as Eq~(\ref{eq:value})
        \ENDIF
    \ENDFOR
    \STATE $e^* \leftarrow \argmax V(s^t, e)$
    \RETURN $\Gamma_\text{H}(e^*)$
\end{algorithmic}
\end{algorithm}

\subsection{Uncertainty-aware Helping Planner}

Finding the optimal helping plan at each step is expensive. To balance the speed and the optimality, our helping planner (Algorithm~\ref{alg:helping_planner}) focuses on finding valuable subgoals instead of updating the whole plan at each step. It considers all edges that appear in the final goal and the intermediate states in the predicted main agent's plans as candidate helping subgoals. Additionally, the helper agent may find that the objects it grabs are no longer needed when it updates goal inference or after the main agent achieves the corresponding subgoals. To allow the helper agent to return those objects to their initial locations, the helping subgoal space also includes edges in the initial state. For each edge $e$, we estimate how long it would take the main agent to reach that subgoal, $L_\text{M}(e)$. If this edge appears in one of the predicted trajectories in the particles, we can conveniently estimate $L_\text{M}(e)$ based on when it appears in the trajectories. If it only appears in the final goals, we then use a fixed length, $L_\text{max}$, to anticipate how many steps it would take the main agent to reach that subgoal. We can also use MCTS to search for a plan for the helper agent, $\Gamma_\text{H}(e)$, to reach the same subgoal. Let $L_\text{H}(e)$ be the length of the helper's plan, we then define the benefit of helping with subgoal $e$ as the speed up the helper agent can offer by reaching the subgoal $e$, i.e., $\max(L_\text{M}(e) - L_\text{H}(e), 0)$. To account for the uncertainty in inference, we estimate how likely $e$ is going to be necessary, $p(e)$, by counting how many particles include $e$ in either the intermediate states or the final goal. Finally, we define a value function for selecting the best subgoal for the helper agent:
\begin{equation}
\begin{array}{ll}
   V(s^t, e) = & w_r p(e) |L_\text{M}(e) - L_\text{H}(e)|_+ - w_c L_\text{H}(e) \\
    & - w_m (\mathcal{D}(s^0, \hat{s}(e)) - \mathcal{D}(s^0, s^t)),
\end{array}
    \label{eq:value}
\end{equation}
where $w_r$, $w_c$, and $w_m$ are constant weights; $\mathcal{D}$ measures the difference between two states; and $\hat{s}(e)$ is the state after reaching the subgoal $e$ from the current state $s^t$. The three terms in Eq.~(\ref{eq:value}) evaluate i) the expected benefit of helping reach the subgoal, ii) the cost of the helper agent, and iii) the additional state change (compared with the initial state) introduced by the subgoal. These three terms make sure that the helper agent selects a subgoal that i) is likely to speed up the task with high certainty, ii) is not too costly for the helper agent, and iii) could restore the initial states of objects that are not needed for the task respectively. Given the subgoal $e^*$, we execute the first action of the helping plan $\Gamma_\text{H}(e^*)$.

\begin{figure*}[t!]
  \centering
\begin{minipage}{.3\textwidth}
    \includegraphics[width=0.98\linewidth]{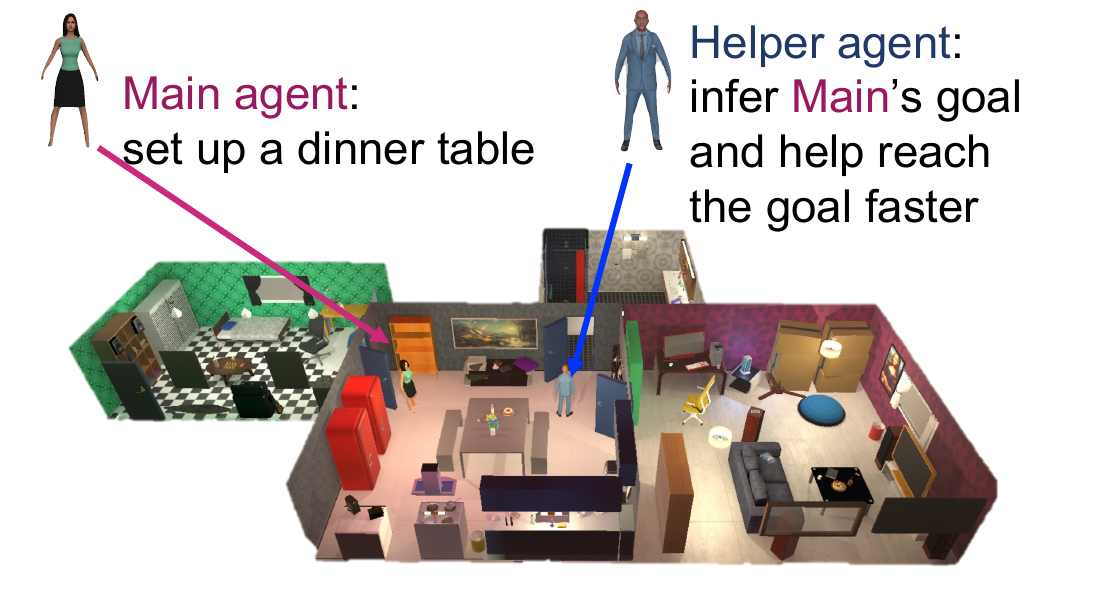}
    \caption{An example setup of O-WAH in one of the simulated apartments.}
    \label{fig:o-wah-overview}
\end{minipage}\hfill
  \begin{minipage}{.68\textwidth}
\scriptsize
    \centering
    \captionof{table}{The goal definition and the number of unique goals for each task type.}
    \begin{tabular}{c|l|c}
         Task Name & Goal definition  & \#Goals  \\ \hline
         \multirow{3}{*}{Set table} & Put  \texttt{N plate}, \texttt{N fork}, \texttt{N OBJ} on \texttt{LOC}, where \texttt{N} $\sim$ \texttt{U(1,3)},  & \multirow{3}{*}{ 12 }  \\
         &   \texttt{OBJ} $\sim$ \texttt{choice([waterglass, wineglass])},\\
         & \texttt{LOC} $\sim$ \texttt{choice([kitchentable, coffeetable])} \\ \hline
         
         \multirow{2}{*}{Put dishwasher} & Put \texttt{N} objects from \texttt{OBJ\_POOL} in \texttt{dishwasher}, where \texttt{N} $\sim$ \texttt{U(3,7)},  & \multirow{2}{*}{315} \\
         & \texttt{OBJ\_POOL = [fork, plates, waterglass, wineglass]} \\ \hline
         
         \multirow{2}{*}{Stock fridge} & Put \texttt{N} objects from \texttt{OBJ\_POOL} in \texttt{fridge}, where \texttt{N} $\sim$ \texttt{U(3,7)},  & \multirow{2}{*}{315} \\
         &  \texttt{OBJ\_POOL = [salmon, apple, cupcake, pudding]}  \\ \hline
         
         \multirow{2}{*}{Prepare meal} & Put  \texttt{N salmon}, \texttt{N apple}, \texttt{N OBJ} on \texttt{LOC}, where \texttt{N} $\sim$ \texttt{U(1,3)},  & \multirow{2}{*}{ 18 }   \\
         & \texttt{OBJ} $\sim$ \texttt{choice([cupcake, pudding])},  \\
         & \texttt{LOC} $\sim$ \texttt{choice([kitchentable, coffeetable, stove])} \\ \hline
         
         \multirow{1}{*}{ Get snacks }  &   Put  \texttt{1 remote}, \texttt{1 condiment}, \texttt{1 chips} on \texttt{coffeetable} & \multirow{1}{*}{ 1 } \\  \hline
         
    \end{tabular}
    
    \label{tab:tasks}
\end{minipage}
\end{figure*}

\begin{figure*}
    \centering
    \includegraphics[width=0.98\linewidth]{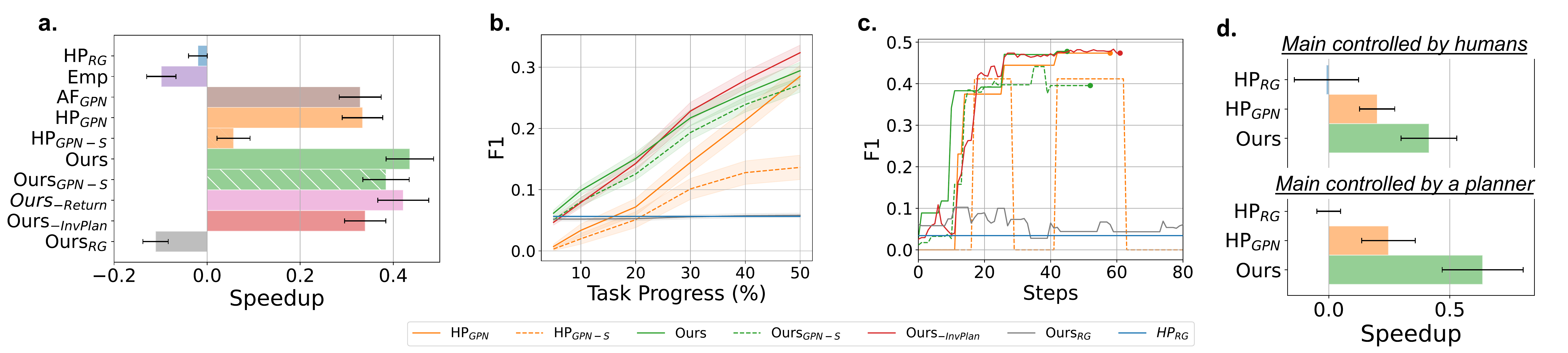}
        \caption{(\textbf{a}) Speedup of different methods (striped bars indicate using the small training set). Errors are standard errors. (\textbf{b}) F1-scores of the predicted goal over the course of a task. The x axis is normalized in proportion to the number of steps needed for the main agent to perform each task alone. The curves show the means and the shaded regions show the standard errors. \textbf{c)} F1-scores over time for different approaches in a single test episode, a dot indicates the number of steps a given baseline took to complete the task. The dashed lines in (\textbf{b}) and (\textbf{c}) indicate using the small training set. (\textbf{d}) Results of the human experiment. Here we show the speedup of different methods when the main agent is controlled by the built-in planner or by human players (note that the results under the two conditions are based on the same 10 testing episodes).}
    \label{fig:result_helper}
\end{figure*}

\section{Online Watch-And-Help}

To evaluate different assistance methods, we propose Online Watch-And-Help (O-WAH), an embodied AI assistance challenge, in which a helper agent has to infer a main agent's goal and help reach the goal as fast as possible. This extends an existing challenge, Watch-And-Help (WAH), to an online assistance problem. O-WAH is built in a realistic multi-agent virtual platform, VirtualHome-Social \cite{puig2021watchandhelp}, simulating daily household tasks (as shown in Fig.~\ref{fig:o-wah-overview}). The goal for each task is defined by a set of predicates and their counts, representing the target locations of different objects in the environment. We sample each goal in the challenge from five general types of household tasks: \textit{set table}, \textit{put dishwasher}, \textit{stock fridge}, \textit{prepare meal}, and \textit{get snacks}. Note that we define task types only to ensure that the goals are emulating real-life household tasks, but that this information is not provided to the helper agents. As summarized in Table~\ref{tab:tasks}, different kinds of uncertainty may arise from these tasks: i) uncertainty in the number of objects, ii) uncertainty in which objects are needed, and iii) uncertainty in the target locations. Compared to prior work, the goal space in O-WAH is 1 or 2 orders of magnitude larger. We adopt the same action space in \cite{puig2021watchandhelp}.  

 \rev{To create a training episode, we first sample a goal and an initial environment using one of the five training apartments and then use a built-in planner to control the main agent to perform the task alone.} The built-in planner is the same hierarchical planner as in \cite{puig2021watchandhelp}. We create a large training set with 6,000 episodes and a small training set with 300 episodes. The testing set has 100 episodes in the two testing apartments unseen during training. 

We use F1-score over the goal predicates to measure the goal inference accuracy. To evaluate the helping performance, we use speedup, where we compare the episode length when the helper agent works with the main agent ($L_\text{H}$) against the episode length when the main agent works alone ($L_\text{M}$), i.e., $L_\text{M} / L_\text{H} - 1$. For each episode, set a time limit of 250 steps and report the average performance across 3 runs.

\section{Experiments}

\subsection{Baselines}

We compare NOPA against several baselines. 

\noindent{$\textbf{HP}_\textbf{GPN}$}: We adopt the best performing approach in the original Watch-And-Help challenge \cite{puig2021watchandhelp} for this baseline, which is a hierarchical planner (HP) based on the most probable goal according to the GPN. In particular, at each step, $\textbf{HP}_\textbf{GPN}$ uses the goal $\hat{g} = \argmax_g q(g | s^0, s^t)$.

\noindent{$\textbf{AF}_\textbf{GPN}$}: We extend $\textbf{HP}_\textbf{GPN}$ by using NOPA's online goal inference. We generate a plan for each predicated goal using HP and execute the most frequent first action among all plans.

\noindent{$\textbf{Empowerment}$}: By adopting the idea of empowerment \cite{du2020ave}, this baseline uniformly samples $K$ goals at each step, predicts plans and intermediate states for the goals, and selects the most frequent edge in the intermediate state as the helping subgoal (i.e., the most common subgoal for \textit{any} goal).

\noindent{$\textbf{HP}_\textbf{RG}$}: A hierarchical planner based on a randomly sampled goal at the beginning of the episode.

We consider the following ablated methods to evaluate the effect of different components of NOPA.

\noindent{$\textbf{Ours}_\textbf{RG}$}: We replace the proposal distribution $q$ in Algorithm~\ref{alg:online_goal_inference} with a uniform distribution.

\noindent{$\textbf{Ours}_\textbf{-InvPlan}$}: Ours without inverse planning.

\noindent{$\textbf{Ours}_\textbf{-Return}$}: Ours without returning \rev{irrelevant} objects to their initial locations ($w_m=0$ in Eq.(\ref{eq:value})).

By default, the GPN is trained on the large training set. To evaluate the sample efficiency of NOPA, we also report the performance of \textbf{Ours} and $\textbf{HP}_\textbf{GPN}$, when the GPN is trained on a small training set, indicated by the subscript \textbf{GPN-S}. \rev{To measure the upper bound on the helping performance, we also implement an oracle helper \noindent{$\textbf{HP}_\textbf{GT}$}, which knows the ground-truth goal and is controlled by an HP.} 

We set $T_\text{max}=250$, $T_\text{prop}=15$, $w_r=1$, $w_c=1$, $w_d=5$, and $L_\text{max}=100$ for NOPA. For all approaches that propose multiple goals, we use $K=20$ proposals. We train the GPN using Adam \cite{kingma2014adam} with a learning rate of $.0009$ and a batch size of 256.

\begin{figure}
    \centering
    \includegraphics[width=1.0\linewidth]{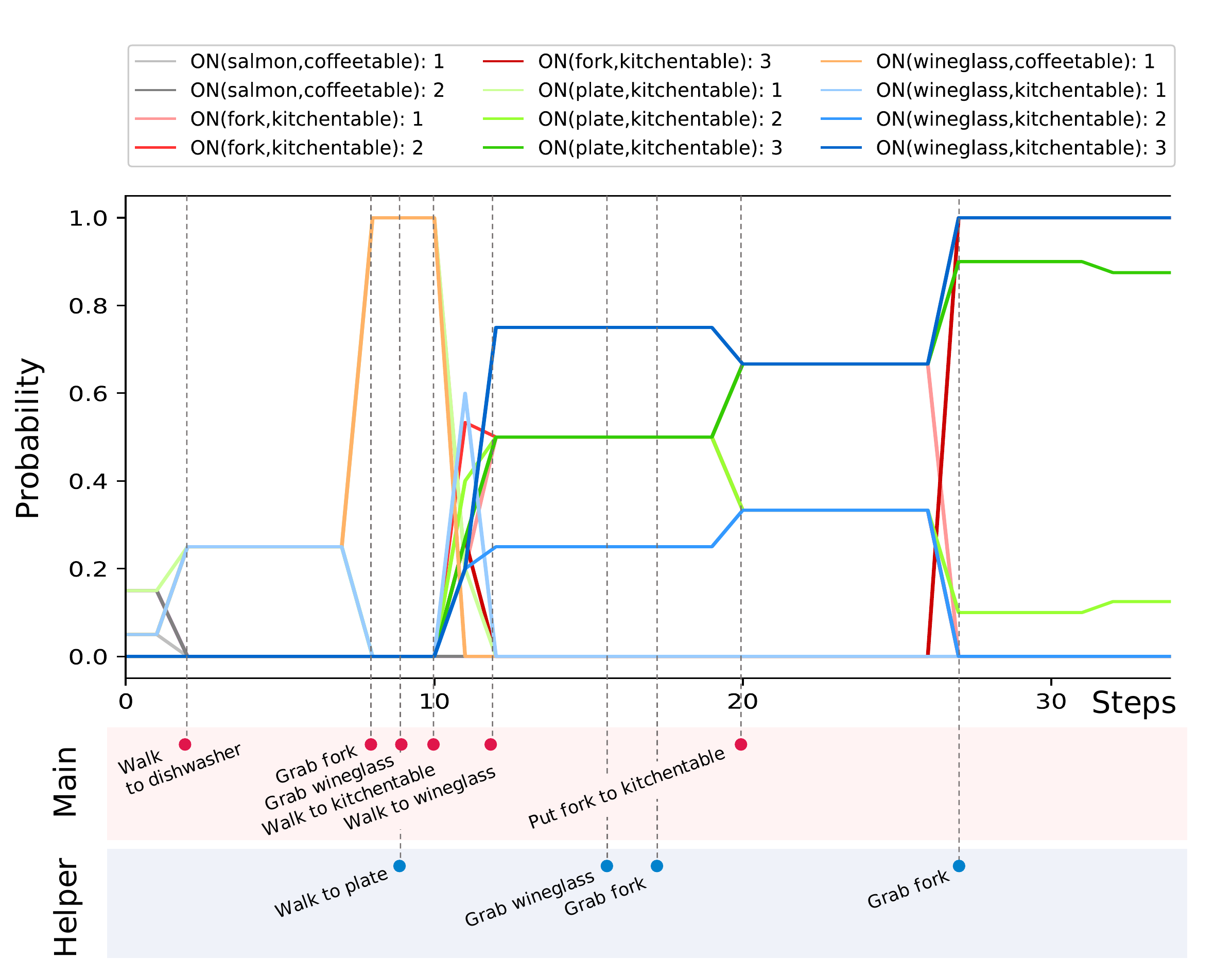}
    \vspace{-10pt}
    \caption{Goal inference and plans by NOPA for the same task shown in Fig.~\ref{fig:result_helper}\textbf{c}, which is setting up a kitchen table for 3 persons. We show the posterior probabilities of the top predicates and their counts based on the particles at each step, key actions of the main agent (indicated by red dots), and key helping actions (indicated by blue dots). At step 2, after watching the main agent walking towards the dishwasher, NOPA rejects proposals involving nearby objects (e.g., apples, salmons) that are not inside of the dishwasher. After the main agent grabs a fork at step 8, NOPA infers with high confidence that the goal is setting up a table for at least one person. So at the following step, the helper agent takes its very first action -- walking to grab a plate. Upon seeing Main walking to the kitchen table at step 10, the goal location becomes certain. \rev{After observing more actions, the inference converges to setting up the kitchen table for 3 persons.}}
    \label{fig:example}
     \vspace{-5pt}
\end{figure}

\begin{figure}[t!]
  \centering
    \includegraphics[width=0.95\linewidth]{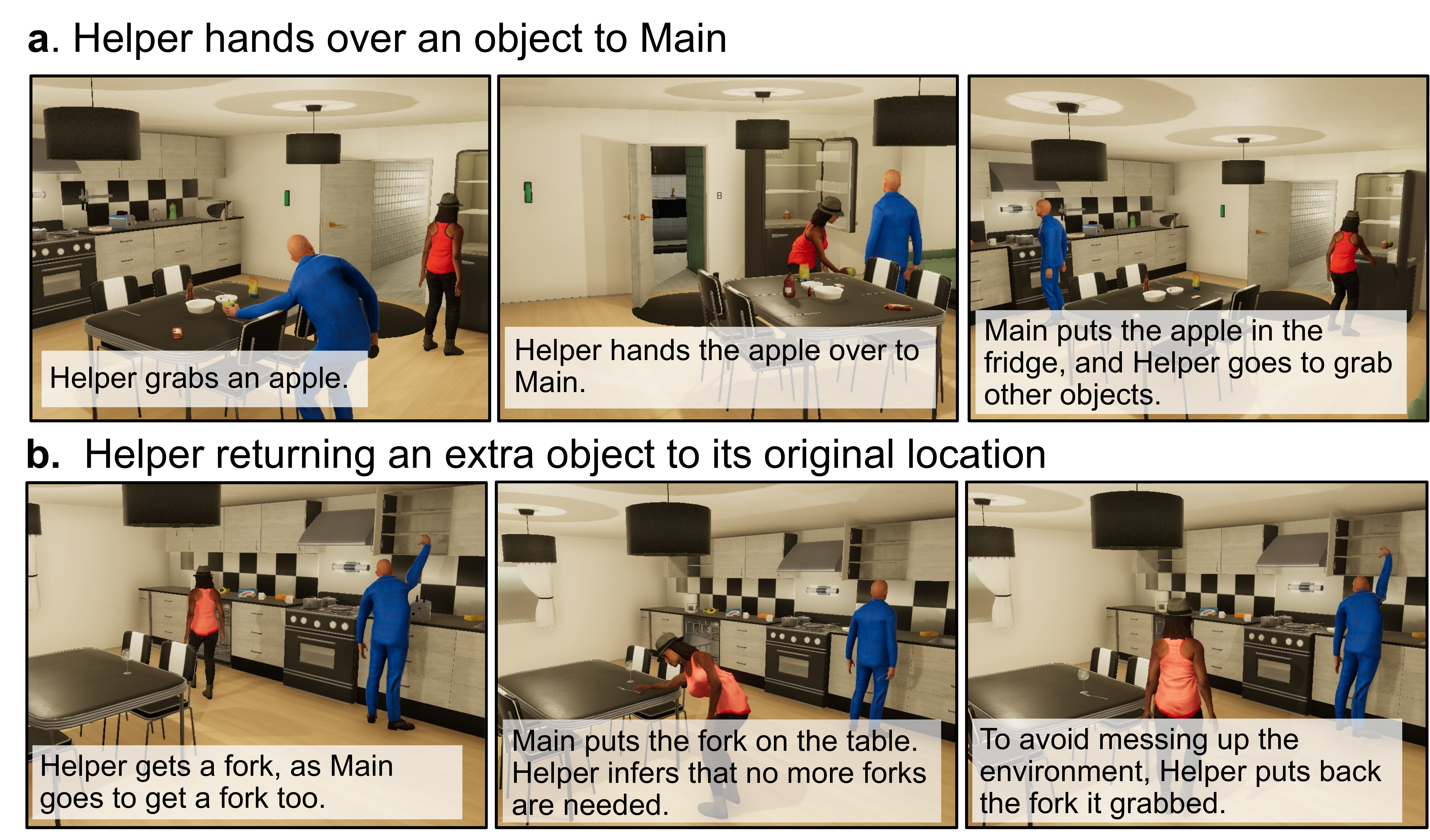}
    \caption{Examples of helping plans that are beyond directly achieving final goals. Main is in red, and Helper is in blue.}
    \label{fig:helping_behaviors}
    \vspace{-5pt}
\end{figure}

\subsection{Results}
\subsubsection{Main Controlled by a Planner}
We evaluate all methods with a main agent controlled by the built-in planner and report the helping speedup (average and standard error across episodes) in Fig.~\ref{fig:result_helper}\textbf{a}. For methods that have different goal inference modules, we also report the F1-score of their goal inference results in Fig.~\ref{fig:result_helper}\textbf{b}. The speedup of the oracle agent, operating with true knowledge about the goal, $\textbf{HP}_\textbf{GT}$ is 1.29. NOPA (\textbf{Ours}) outperforms all baselines, offering the highest speedup. It also achieves the best goal inference accuracy at the early stage of the tasks, which serves as the foundation of its successful assistance. This benefit can be more clearly seen from Fig.~\ref{fig:result_helper}\textbf{c} (the improvement margin appears to be smaller since the temporal normalization for each episode is different). The low speedup by \textbf{Empowerment} suggests that online goal inference is necessary for effective assistance, despite its success in certain domains shown in prior work \cite{du2020ave}. Given that the predicted goal may be uncertain, using multiple goal proposals leads to a better helping performance, as seen by comparing $\textbf{Ours}_\textbf{GPN}$ with $\textbf{HP}_\textbf{GPN}$. The effect is more pronounced when the GPN is trained with fewer data and is consequently less accurate ($\textbf{GPN-S}$). We also find that the \rev{neurally-guided} goal proposals can greatly improve the goal inference over uniform goal proposals ($\textbf{Ours}_\textbf{RG}$). Moreover, the results demonstrate that inverse planning is important for filtering spurious goal proposals from $\textbf{GPN}$, significantly improving the speedup over $\textbf{Ours}_\textbf{-InvPlan}$ since it allows the goal inference to reach a relatively high accuracy much earlier than $\textbf{Ours}_\textbf{-InvPlan}$ and other baselines do (see Fig.~\ref{fig:result_helper}\textbf{c}). Finally, by comparing NOPA with $\textbf{Ours}_\textbf{-Return}$, we can see a marginal improvement in speedup by avoiding unnecessarily distorting the environment; $\textbf{Ours}_\textbf{-Return}$ also causes 11.2\% more unnecessary state changes.

Fig.~\ref{fig:example} shows a typical successful example by NOPA, where the task is the same as the one in Fig.~\ref{fig:result_helper}\textbf{c}. It demonstrates that NOPA can i) achieve accurate goal inference early on by filtering out goal proposals that are inconsistent with the main agent's actions, ii) correctly update the goal inference and its uncertainty estimation based on more observation, and iii) plan for effective helping actions based on the filtered goal proposals and the uncertainty in the inference. Note that the helper remains idle but takes a useful helping action as soon as the goal inference becomes confident and is able to avoid grabbing extra objects thanks to its gradual update of the number of objects needed.

We also observe diverse helping behaviors enabled by NOPA that are not just about directly achieving the final goals as shown in  Fig.~\ref{fig:helping_behaviors}. First, the helper agent sometimes selects a subgoal of handing over objects to the main agent. For example, as illustrated in Fig.~\ref{fig:helping_behaviors}\textbf{a}, the helper agent hands over the apple to the main agent who is right next to the fridge so that the task execution can be faster. Second, the helper agent can return extra objects to their initial locations once it realizes that they are not needed for reaching the goal (Fig.~\ref{fig:helping_behaviors}\textbf{b}). The supplementary video\footnote{The supplementary video is available at https://youtu.be/Oawo9pynPL0.} shows more examples.

\subsubsection{Main Controlled by Humans}

To evaluate how effective helper agents are at assisting real humans, we conducted a human experiment where the main agent is controlled by human players. We used 10 testing episodes to run 40 trials. In each trial, a human participant was asked to either perform the task alone (to estimate the number of steps needed for completing each task alone) or work with a helper agent controlled by one of the three approaches, NOPA, $\textbf{HP}_\textbf{GPN}$, and $\textbf{HP}_\textbf{RG}$. Participants did not know which helper agent they were working with. We recruited 10 participants (mean age = 32.3; 4 female) who had no prior exposure to our system. As shown in Fig.~\ref{fig:result_helper}(\textbf{d}), the ranking of the methods remains consistent when the main agent is controlled by human players. There is no significant difference in NOPA's performance under the two conditions ($t(9)=0.87$, $\rho= 0.40$).

\section{Conclusion}

In this work, we propose a novel method for building socially intelligent home assistants, Neurally-guided Online Probabilistic Assistance (NOPA), which integrates (1) a hybrid online goal inference algorithm combining a goal proposal network and inverse planning and (2) an uncertainty-aware helping planner that \rev{identifies} valuable helping subgoals from both the final goals and intermediate states. For a systematic and scalable evaluation, we introduce a new embodied AI assistance challenge, Online Watch-And-Help, based on a realistic virtual home platform. We evaluate NOPA with several baselines in our challenge with a main agent controlled either by a built-in planner or by humans. Our experiments show that NOPA significantly outperforms baselines and achieves great sample efficiency for training.  In the future, we plan to extend NOPA to a partial observability setting and apply it to robots in real homes.

\section*{ACKNOWLEDGMENT}
This work was supported by the DARPA Machine Common Sense program, ONR MURI N00014-13-1-0333, and a grant from Lockheed Martin.

\bibliographystyle{IEEEtran}
\bibliography{IEEEabrv,IEEEfull}

\end{document}